\documentclass[DIV=14, letterpaper]{scrartcl}

\usepackage{amsmath,amssymb,amsfonts,amsthm}
\usepackage{graphicx}
\usepackage{mathptmx}
\usepackage{cite}
\usepackage{hyperref}
\usepackage{subcaption}
\usepackage{float}
\usepackage{url}

\usepackage{tikz}
\usetikzlibrary{calc}
\usetikzlibrary{matrix}
\usepackage{esvect}

\def \p{\partial}
\def \DD{\mathbb{D}}
\def \DDK{\mathbb{D}_K}
\def \MM{\mathbb{M}}
\newcommand{\bey}{\begin{eqnarray}}
\newcommand{\eey}{\end{eqnarray}}

\newcommand{\beq}{\begin{equation}}
\newcommand{\eeq}{\end{equation}}
\theoremstyle{plain}

\theoremstyle{definition}

\theoremstyle{remark}
\newtheorem{exam}{\hspace{6mm}\textbf{Example}}[section]

\begin{document}

\title{Anisotropic Mesh Adaptation for Image Segmentation Based on Mumford-Shah Functional}
\author{Karrar Abbas 
\thanks{
Department of Mathematics and Statistics, the University of Missouri-Kansas City, Kansas City, MO 64110,
U.S.A. (\textit{kka8cc@mail.umkc.edu})}
\and Xianping Li 
\thanks{
College of Integrative Sciences and Arts, Arizona State University, Mesa, AZ 85212,
U.S.A. (\textit{Xianping.Li@asu.edu})}
}

\date{}

\maketitle

\textbf{Abstract}
\begin{abstract}
	As the resolution of digital images increase significantly, the processing of images becomes more challenging in terms of accuracy and efficiency. In this paper, we consider image segmentation by solving a partial differentiation equation (PDE) model based on the Mumford-Shah functional. We develop a new algorithm by combining anisotropic mesh adaptation for image representation and finite element method for solving the PDE model. Comparing to traditional algorithms solved by finite difference method, our algorithm provides faster and better results without the need to resizing the images to lower quality. We also extend the algorithm to segment images with multiple regions.	  
\end{abstract}


\section{Introduction}
\label{Sec-intro}

In computer vision, image segmentation is the process of segmenting the image into different regions (sets of pixels/voxels). In other words, it is the process of information detection and extraction from image data for certain regions that share similar features throughout the image \cite{shapirocomputer,kang2009comparative}. The result of this process is either a set of segments that all together cover the whole image, contours extract of the image, or separate objects that are saved to separated image files. During this process the image is classified to different sets of pixels/voxels belonging to two categories: similar and adjacent regions. The pixels/voxels that belong to similar region contain similar properties with respect to texture, intensity or color while adjacent regions are significantly different with respect to those characteristics.

The goal of image segmentation is to locate a certain object with similar pixels/voxels, and simplify the image so it is easier to analyze \cite{barghout2004perceptual}. Image segmentation is also used to locate the edges or boundaries of certain objects of an image \cite{kang2009comparative}. There are many important applications for image segmentation such as medical imaging (locating tumors, measuring tissue volumes, surgery planning) \cite{forouzanfar2010parameter, george2012mr}, object detection (pedestrian detection, face detection, satellite image detection) \cite{delmerico2011building}, recognition tasks (face, finger, and iris recognition), and traffic control system (video surveillance) \cite{wang2018segment}. 

Various methods and algorithms have been developed for image segmentation that can be categorized into two groups: local segmentation and global segmentation. Local segmentation deals with segmenting the pixels/voxels of specific parts of the images while global segmentation deals with the entire image \cite{khan2011image, janowczyk2012high}. Segmentation methods can also be classified according to their approaches, including region approach, edge approach, and boundary approach. Region approach such as region growing method performs generally better at noise resistance than edge detection method, but it is very costly when it comes to computation \cite{kang2009comparative}. Boundary or edge approach such as edge detection method works well with images containing high regional and boundary disparity, but not with images containing blurred edges or noises \cite{zhang2006overview}. One of the most well-known and commonly used method worth mentioning is the thresholding method that does not require formal knowledge of the images. Segmentation is applied based on the intensity level of image pixels. This method is fast, simple to implement, and has less computational cost. However, it only works well with images that have  good background to foreground contrast, so the segmentation may lack object coherency (i.e. may have holes or irrelevant pixels) \cite{kang2009comparative, zhang2006overview}. Other methods include structural method that requires a formal knowledge, stochastic method that is based on pixel values, and hybrid method that combines both features of structural and stochastic methods \cite{anjna2017review,wahba2009automated}. 

In recent years, partial differential equation models (PDEs) and variational methods have been widely applied to image segmentation due to their well-established mathematical fundations. The main idea is to find the minimization cost of functionals by evolving a curve to approach the lowest potential cost of the function. Mumford-Shah functional \cite{mumford1989optimal} is one of the commonly used variational methods. Although many mathematical theories have been developed, there is still a lack of efficient computational methods. For example, the Chan-Vese algorithm is one of the well-known algorithms used to solve a simplified model based on the Mumford-Shah functional \cite{evans1998graduate} for segmentation. It needs many iterations before converging to its final solution \cite{getreuer2012chan}. On the other hand, some models that are easier to solve numerically encounter other difficulties. For example, active contour models can be greatly influenced by noise, hence Gaussian smoothing methods are needed to decrease the effect of noise. However, the boundaries will also be smoothed, which makes it difficult to distinguish the edges. Perona-Malik anisotropic diffusion method was proposed to smooth the pixels inside the regions rather than on the boundaries \cite{perona1990scale}. Although the method performs well in general, the mathematical problem is ill-posed in the sense that a weak solution is not guaranteed \cite{kichenassamy1997perona}. 

The goal for this paper is to develop a computational method for image segmentation that improves the computational efficiency comparing to the Chan-Vese algorithm as well as reduces the effect of noise. We will focus on two-dimensional images in this paper. Application to three-dimensional images can follow the similar procedures. 

Our method is based on the anisotropic mesh adaptation (AMA) method. We firstly represent the image using an anisotropic triangular mesh with fewer points than the pixels in the original image \cite{li2016anisotropic} and denote the representation as the AMA image. Secondly, we solve the PDE segmentation model \eqref{pdemodel} using finite element method with the AMA image as the input. Finally, we reconstruct the segmented image from the finite element solution obtained on the anisotropic triangular mesh. By applying the AMA method, segmentation of high resolution images can be performed much faster than the classic Chan-Vese algorithm, without rescaling the images to lower qualities. 

The outline of this paper is as follows. Section \ref{Sec-mumford} is an introduction to the Mumford-Shah Functional as well as the Chan-Vese model. Section \ref{Sec-amaseg} describes our AMA image segmentation framework. Some numerical results are presented in Section \ref{Sec-results} and some conclusions are drawn in Section \ref{Sec-con}.

\section{Mumford-Shah Functional and Chan-Vese model}
\label{Sec-mumford}

In this section, we briefly describe the Mumford-Shah functional and the Chan-Vese model. More details can be found in \cite{mumford1989optimal,getreuer2012chan}.
Mumford-Shah functional is one of the most famous mathematical approaches in image processing and is the mathematical foundation for the well-known Chan-Vese model. Mumford-Shah functional is a differential geometric technique that treats the image as a compact space for the process of a piecewise smooth segmentation. The minimization of the functional is formulated as follows. 
\beq
\label{mumford-shah}
\arg {\min _{u,C}} \quad \mu \cdot  Length(C) + \lambda \int\limits_\Omega  \left( (f(x) - u(X) \right)^2 dx + \int\limits_{\Omega /C} \left| {\nabla u(X)} \right|^2 \, dx,
\eeq
where $f$ is a grayscale image on a domain $\Omega \subset \mathbb{R}^2$, $C$ is an edge curve set, $\mu$ and $\lambda$ are positive parameters, and $u:\Omega  \to \mathbb{R}$ is the segmented solution that is allowed to be discontinuous on $\Omega$. The first term in \eqref{mumford-shah} ensures regularity of boundary curve $C$, the second term enforces $u$ to be close to $f$, and the third term makes sure that $u$ is differentiable on $\Omega /C$.

In practice, a simplified version of the Mumford-Shah model is usually considered
\beq
\label{mumford-shah-s}
\arg {\min _{u,C}} \quad  \mu \cdot Length(C) + \lambda \int\limits_\Omega  \left( (f(x) - u(X) \right)^2 dx,
\eeq
where $C$ is a closed set and $u$ is required to be constant on each connected component of $\Omega/C$. The above formula is considered as a piecewise constant formulation. The existence of solution for \eqref{mumford-shah-s} is proved by Mumford and Shah \cite{mumford1989optimal}. 

Various methods have been developed to solve the minimization problem \eqref{mumford-shah-s}, such as variational method \cite{aubert2006mathematical}, elliptic approximation method \cite{ambrosio1990approximation}, and curve evolution method \cite{chan2001active,gao2005image}. The curve evolution method is shown to be stable \cite{gage1986heat,sethian1999level}, with the well-known Chan-Vese model as a particular example \cite{getreuer2012chan}.
Chan-Vese model further simplifies the Mumford-Shah functional by allowing $u$ to take only two values and adding a penalty term using the enclosed area. The value of $u$ is represented as follows.
\beq
\label{uvalue}
u( x ) = \left\{ \begin{gathered}
  c_1,\text{ where }x{\text{ is inside }}C \hfill \\
  c_2,\text{where }x{\text{ is outside }}C \hfill \\ 
\end{gathered}  \right.\ ,
\eeq
where $C$ is the boundary of a closed set and $c_1$, $c_2$ are the values of $u$ inside and outside
of $C$, respectively. The Chan-Vese model is formulated as follows.
\beq
\label{chan-vese}
\arg {\min _{u,C}} \quad  \mu \cdot Length(C) + \nu \cdot Area(inside(C)) + \; \lambda_1 \int\limits_{inside(C)} |f(x) - c_1|^2 dx
  + \; \lambda_2 \int\limits_{outside(C)} |f(x) - c_2|^2 dx.
\eeq
The first term in \eqref{chan-vese} controls the curve length, the second term controls the size of the enclosed area $C$. The third and fourth terms in \eqref{chan-vese} control the difference between the piecewise constant model $u$ and the image value $f$. The segmentation is obtained by finding a local minimizer of the Chan-Vese model \eqref{chan-vese}. The existence of solution for \eqref{chan-vese} is proved in \cite{morel2012variational}. 

\begin{figure}[ht!]
\centering
\includegraphics[width=2.5in]{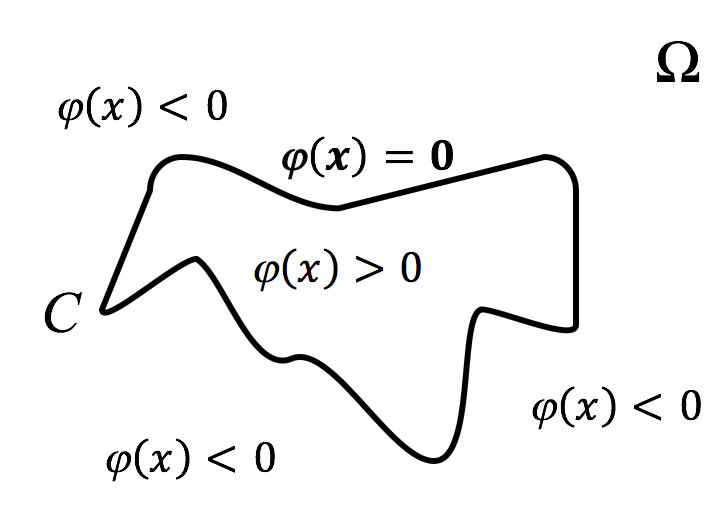}
\caption{Level set function for a curve $C$ inside domain $\Omega$.}
\label{Levelset}
\end{figure}

To solve the Chan-Vese model \eqref{chan-vese}, we use one of the curve evolution methods - level set method  \cite{getreuer2012chan,sumengen2002image }. Here we introduce the level set function $\varphi(x)$ to represent the enclosed curve $C$ as follows \cite{osher1988fronts}. A sketch of the representation is shown in Figure \ref{Levelset}.
\begin{equation}
\label{curveC}
\begin{gathered}
C = \left\{x\in{\Omega} \subset {\mathbb{R}^2}:\varphi(x)=0 \right\} 
\\
\text{inside curve } C = \left\{x\in{\Omega} \subset {\mathbb{R}^2}:\varphi(x)>0 \right\} 
\\
\text{outside curve } C = \left\{x\in{\Omega} \subset {\mathbb{R}^2}:\varphi(x)<0 \right\} 
\end{gathered}.
\end{equation}

Representing $C$ by the level set function is beneficial because level set method adapts better to automatic topological changes and allows for cusp corners. Any level set function that satisfies \eqref{curveC} (or similar) can be consider as a valid representation for the curve $C$. 

Let $H$ and $\delta$ denote the Heaviside function and its derivative, receptively, and use $H(\varphi)$ as the indicator of the level set function as follows.
\begin{align}
\label{Hvalue}
H(\varphi) = \left\{ \begin{gathered} 
1 \quad \text{ where } \varphi \geq 0      
\\                             
0 \quad \text{ where } \varphi < 0        
\end{gathered} \right. ,  \qquad \delta  = \frac{d}{d\varphi} H(\varphi).     
\end{align}
Then the first two terms of \eqref{chan-vese} can be written as
\begin{equation}
\begin{gathered}
Length(C)=Length(\varphi=0)=\int \limits_\Omega {\delta(x) |\nabla \varphi(x)|} dx, \\
Area(\varphi\geq 0)=\int \limits_\Omega {H( \varphi(x))} dx. 
\end{gathered}
\end{equation}
Hence Chan-Vese model \eqref{chan-vese} can be rewritten as follows.
\begin{align}
\label{chan-vese-2}
\begin{gathered}
\arg {\min_{c_1,c_2,\varphi}} \quad  \mu \int \limits_\Omega {\delta(x) |\nabla \varphi(x)|} dx + \nu \int \limits_\Omega {H( \varphi(x))} dx + \lambda_1 \int\limits_{inside(C)} |f(x) - c_1|^2 H( \varphi(x)) dx \\
+  \lambda_2 \int\limits_{outside(C)} |f(x) - c_2|^2  \left(1 - H( \varphi(x)) \right) dx,
\end{gathered}
\end{align}
where $c_1$ and $c_2$ are chosen as the region averages
\begin{equation}
\label{value-c1,c2}
c_1 = \frac{\int f(x) H(\varphi(x)) dx}{\int  H(\varphi(x)) dx}, \quad c_2 = \frac{\int f(x) (1-H(\varphi(x))) dx}{\int (1-H(\varphi(x))) dx}.
\end{equation}

In our computations, $H(\varphi)$ is regularized as
\begin{equation}
\label{Hvalue2}
H(\varphi)=\frac{1}{2} \left(1+\frac{2}{\pi} \arctan \left(\frac{\varphi}{\epsilon} \right) \right),  
\end{equation}
for some parameter $\epsilon>0$. Hence, we have

\begin{equation}
\label{Deltavalue}
\delta_\epsilon(\varphi)= \frac{d}{d\varphi} H = \frac{\epsilon}{\pi (\epsilon^2+\varphi^2)}.
\end{equation}

Keeping both $c_1$ and $c_2$ fixed, the solution of the Chan-Vese model \eqref{chan-vese-2} can be obtained by solving its Euler-Lagrange equation
\bey
\label{pdemodel}
&\frac{\p\varphi}{\p t}= \delta_{\epsilon}(\varphi) \left[ \mu \; \nabla \cdot \left(\frac{\nabla\varphi}{|\nabla\varphi|}\right) -\, \nu \, - \,\lambda_1 \,(f(x)- c_1)^2 + \, \lambda_2 \, (f(x)- c_2)^2 \right] \qquad \text{ in }\  \Omega 
\\
\label{pdebc}
&\frac{\delta_\epsilon(\varphi) }{|\nabla\varphi|}\frac{\p\varphi}{\p \vv{n}}=0 \qquad \text{ on } \ \p\Omega,
\eey
where $t$ is an artificial time parameterizing for the descent direction, and $\vv{n}$ is the outward normal on the image boundary.

\section{AMA image segmentation framework}
\label{Sec-amaseg}

In this section, we describe the AMA image segmentation framework, including AMA method for image representation, the finite element method (FEM) in numerical computation, and our algorithm for image segmentation. 

\subsection{AMA representation}
\label{Sec-amaM}

Anisotropic mesh adaptation (AMA) has been successfully applied in numerical computations to improve the efficiency and accuracy \cite{FA05, HL08, li2010anisotropic, DZ10, LH13, WDLQV12, Li18}. Recently, Li applied AMA for image representation \cite{li2016anisotropic} and developed GPRAMA method that provides better representation quality than other methods of comparable computational cost. In this subsection we briefly describe the AMA representation method that will be used in our computations. More details can be found in \cite{huang2006mathematical,li2010anisotropic,li2016anisotropic}.

Our AMA method takes the $\MM$-uniform mesh approach while the mesh is generated based on a metric tensor $\MM=\MM(x)$ that is required to be symmetric and positive definite. The metric tensor $\MM$ provides specific information related to mesh elements include size, shape, and orientation necessary to the mesh generation \cite{huang2006mathematical}. The anisotropic physical mesh is viewed as a uniform mesh in the metric specified by $\MM$.

Let $\Omega$ be a 2D domain and $\mathcal{T}_h$ be a triangular mesh in $\Omega$. Denote $F_K: \hat{K}$ $\to$ $K$ as the affine mapping from the reference element $\hat{K}$ to a triangular element $K \in \mathcal{T}_h$. The reference element $\hat{K}$ is chosen to be equilateral and unitary in area. It is shown that an $\MM$-uniform 2D triangular mesh $\mathcal{T}_h$ generated by metric tensor
$\MM=\MM(x)$ satisfies the following two conditions \cite{huang2006mathematical}
\bey
\label{cond-eq}
&& |K| \sqrt{\mbox{det}(\MM_K)} =  \frac{\sigma_h}{N}, \quad \forall K \in \mathcal{T}_h, \\
\label{cond-ali}
&& \frac{1}{2} \mbox{tr} \left ( (F_K')^T \MM_K F_K' \right ) = 
\mbox{det} \left ( (F_K')^T \MM_K F_K' \right )^{\frac{1}{2}}, \quad \forall K \in \mathcal{T}_h,
\eey
where $F_K'$ is the Jacobian matrix of $F_K$, $|K|$ is the area of the element $K$, $N$ is the number of mesh elements, and
\begin{equation}
\MM_K = \frac{1}{|K|} \int_K \MM(\mathbf{x}) d \mathbf{x}, \qquad 
\sigma_h = \sum_{K \in \mathcal{T}_h} |K| \sqrt{\mbox{det}(\MM_K)}.
\end{equation}
Conditions (\ref{cond-eq}) and (\ref{cond-ali}) are called the {\em equidistribution condition} and {\em alignment condition}, respectively. Equidistribution condition determines the size of element $K$, and alignment condition regulates the shape and orientation of $K$.

There are different formulations for metric tensors. In this paper, we choose two forms of metric tensors for AMA representation. One is developed in \cite{huang2005metric} and denoted as $\MM_{anios}$, the other is $\MM_{DMP}$ developed in \cite{li2010anisotropic}. Let $H_K$ denote the value of the Hessian matrix $H$ at the center of element $K$, $\|\cdot\|_F$ be the Frobenius matrix norm. 
The metric tensor $\MM_{anios,K}$ is defined over a triangular element $K$ as follows
\beq
\label{M-aniso}
\MM_{aniso,K}=\rho_K \det
\left( I+\frac{1}{\alpha _{h}}|H_K|\right) ^{-\frac{1}{2}}
\left[ I+\frac{1}{\alpha _{h}}|H_K|\right] ,
\eeq
where $\rho_K$ is the density function defined as
\beq
\rho_K = \Big \| I + \frac{1}{\alpha_h} | H_K | \Big \|_F ^{\frac{1}{2}
}\,\det \left( I+\frac{1}{\alpha _{h}}|H_K|\right)^{\frac{1}{4}},
\eeq
$I$ is the $2 \times 2$ identity matrix, $\alpha_h$ is the regularization parameter and is defined implicitly through
\beq
\sum_{K\in \mathcal{T}_h} \rho_K |K| = 2 | \Omega | .
\eeq
With this choice of $\alpha_h$, roughly fifty percents of the triangular elements will be concentrated in large gradient regions \cite{huang2005metric}.

For our other choice of metric tensor, we consider a general anisotropic diffusion equation in the form of 
\beq
\label{eq-aniso}
u_t - \nabla \cdot (\DD \nabla u) = g(u,x,t) \qquad \mbox{ in } \Omega
\eeq
with $\DD=\DD(x)$ being the diffusion matrix that is symmetric and strictly positive definite on $\Omega$. In this paper, we choose $\DD$ proposed in \cite{wang2016image} as follows
\beq
\label{DD_eig}
\DD = \frac{1}{r(r-1)} \begin{bmatrix} u_x^2 + r u_y^2 & (1-r)u_x u_y \\ (1-r)u_x u_y & u_y^2 + r u_x^2 \end{bmatrix},
\qquad \text{ with } r=1+(\nabla u)^T \nabla u.
\eeq
Then we define $\MM_{DMP}$ over $K$ as 
\beq
\MM_{DMP,K} = \DDK^{-1},\quad \forall K \in \mathbb{T}_h
\eeq
with 
\beq
\DDK=\dfrac{1}{K} \ \int \limits_K \DD(x) \,dx.
\eeq
The mesh generated according to $\MM_{DMP}$ is called a DMP mesh. The linear finite element solution for \eqref{eq-aniso} using a DMP mesh is guaranteed to satisfy the discrete maximum principle (DMP) \cite{li2010anisotropic}.  

Once the metric tensor $\MM$ is computed on the initial or current mesh, the free C++ code BAMG (bidimensional anisotropic mesh generator) \cite{hecht2010bidimensional} is utilized to generate the triangular mesh according to $\MM$. Results obtained using different metric tensors are presented in Section \ref{Sec-results}.

\subsection{Finite Element Method}

In this section, we consider the linear finite element solution for the PDE in \eqref{pdemodel} with boundary condition \eqref{pdebc}. Denote
\beq
\label{defD}
\DD = \frac{\mu}{|\nabla\varphi|}, \mbox{ and } F(\varphi)= -\, \nu \, - \,\lambda_1 \,(f(x)- c_1)^2 + \, \lambda_2 \, (f(x)- c_2)^2.
\eeq
Then equation \eqref{pdemodel} can be rewritten as
\beq
\label{pdemodel-2}
\frac{\p \varphi}{\p t} - \delta_{\epsilon}(\varphi) \nabla \cdot (\DD \nabla \varphi) = \delta_{\epsilon}(\varphi) F(\varphi).
\eeq

Let $\Omega \subset \mathbb{R}^2$ be a connected polygon which has the same size of the image dimension. Let $\left\{ \mathcal{T}_h \right\}$ be an affine family of simplicial triangulations given on $\Omega$. Denote $U_0 =\left\{v \in H^1(\Omega) : v|_{\p\Omega}=0 \right\}$ and let $U_0^h \subset U_0$ be a linear finite element space associated with mesh $\mathcal{T}_h$. Then a linear finite element solution $\varphi^h(t) \in U_0^h$ of \eqref{pdemodel-2} is given as follows
\begin{equation}
\label{weakform}
\int_\Omega \frac{\p \varphi^h}{\p t} v^h \, dx\, + \int_\Omega \delta_\epsilon(\varphi^h)(\nabla v^h)^T \mathbb{D}\, \nabla \varphi^h \, dx =\int_\Omega \delta_\epsilon(\varphi^h)\,F(\varphi^h) \,v^h \, dx,\quad \forall v^h\in U_0^h, \, t \in (0,T].
\end{equation}
Denote the number of vertices of $\mathcal{T}_h$  by $N_v$, and let $\phi_j$ be the linear basis function associated with vertex $x_j$. Then we can express the solution $\varphi^h$ as
\begin{equation}
\label{basis}
\varphi^h= \sum_{j=1}^{N_v} \varphi_j(t)\phi_j.
\end{equation}

Substituting (\ref{basis}) into (\ref{weakform}) and taking $v^h =\phi_i \, (i=1,...,N_v)$, we obtain the linear system
\begin{equation}
\label{FEM-EQ}
M \frac{d\vec{\varphi}}{dt}+A(\varphi^h)\vec{\varphi}=\vec{b}(\varphi^h),
\end{equation}
where $\vec{\varphi}=(\varphi_1,...,\varphi_{N_v})^T$ is the unknown vector, $M$ and $A$ are the mass and stiffness matrices, respectively, and $\vec{b}$ is the right-hand side vector. The entries of $M$, $A$ and $\vec{b}$ are given as follows
\bey
&&m_{ij}=\int_\Omega \phi_j\phi_i \,dx, \qquad i,j=1, \cdots, N_v \\
&&a_{ij}=\int_\Omega \delta_\epsilon (\varphi^h)(\nabla\phi_i)^T\mathbb{D}\nabla\phi_j \,dx, \qquad i,j=1, \cdots, N_v\\
&&b_i=\int_\Omega \delta_\epsilon (\varphi^h)F(\varphi^h)\phi_i \,dx, \qquad i=1, \cdots, N_v. 
\eey

Regarding the time discretization, we denote the numerical solution at $t=t_n$ by $\vec{\varphi}^n$. Applying the semi-implicit scheme to \eqref{FEM-EQ}, we get
\begin{equation}
\label{Bc-Euler}
M \frac{\vec{\varphi}^{n+1}-\vec{\varphi}^n}{\Delta t_n}+\tilde{A} \cdot \vec{\varphi}^{n+1}=\tilde{\vec{b}},
\end{equation}
where $\Delta t_n \,=t_{n+1}-t_{n}$, and $\tilde{A}$ and $\tilde{\vec{b}}$ are approximations of $A$ and $\vec{b}$ at $t_{n}$.

\subsection{AMA segmentation algorithm}

\begin{figure}[tbh]
\centering
\tikzset{my node/.code=\ifpgfmatrix\else\tikzset{matrix of nodes}\fi}
\begin{tikzpicture}[every node/.style={my node},scale=0.45]
\draw[thick] (0,0) rectangle (6,3.5);
\draw[thick] (8,0) rectangle (14,3.5);
\draw[thick] (16,0) rectangle (22,3.5);
\draw[thick] (24,0) rectangle (30,3.5);
\draw[thick] (24,-1) rectangle (30,-4.5);
\draw[thick] (16,-1) rectangle (22,-4.5);
\draw[->,thick] (6,1.75)--(8,1.75);
\draw[->,thick] (14,1.75)--(16,1.75);
\draw[->,thick] (22,1.75)--(24,1.75);
\draw[->,thick] (27,0)--(27,-0.5)--(23,-0.5)--(23,-2.5)--(22,-2.5);
\draw[->,thick] (30,1.75)--(31,1.75)--(31,-2.5)--(30,-2.5);
\node (node1) at (3,1.75) {Given an image\\};
\node (node2) at (11,1.75) {Represent the \\ image using \\ anisotropic mesh\\};
\node (node3) at (19,1.75) {Solve the PDE \\ model \\};
\node (node4) at (27,1.75) {Reconstruct \\ using FEM \\ interpolation\\};
\node (node5) at (27,-2.75) {Displaying \\ evolving \\ curve $\varphi$\\};
\node (node6) at (19,-2.75) {Displaying \\ segmented \\ regions\\};
\end{tikzpicture}
\caption{A sketch of the procedures for image segmentation using anisotropic mesh adaptation.}
\label{fig-procedure}
\end{figure}
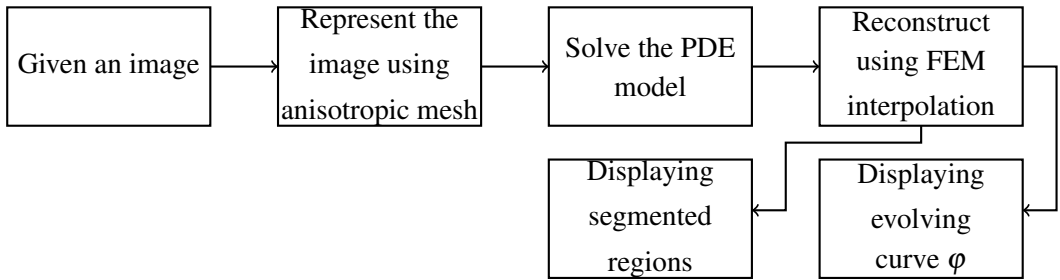

Our AMA segmentation algorithm consists of three main stages: representing the image using an adaptive mesh, solving the PDE model using FEM, and reconstruct the numerical solution back to an image. Details of the first and third stages are described in \cite{li2016anisotropic}. The basic idea is to compute a metric tensor according to the image gray values and generate an anisotropic mesh for the image, then finite element interpolation is used to reconstruct the segmented image or evolving curve. Here, we focus on the second stage. Firstly, an initial condition $\vec{\varphi}^o(x)$ is chosen as follows.
\begin{equation}
\label{initial-cond}
\vec{\varphi}^o(x)= \sin(\frac{\pi}{4}x_1).\sin(\frac{\pi}{4}x_2)
\end{equation}
Then, the values of the parameters are computed such as the Heviside function $H$ as in (\ref{Hvalue2}), its derivative $\delta$ as in (\ref{Deltavalue}), and $c_1,\, c_2$ as in (\ref{value-c1,c2}). Finally, the linear system (\ref{Bc-Euler}) is assembled and solved. The sketch of the procedures is shown in Figure \ref{fig-procedure}. Note that anisotropic mesh adaptation is only performed for AMA representation and the corresponding mesh is used when solving the PDE model with finite element method. 

The above procedures work for an image with one region of interest, i.e, containing only one object and background. For images containing multi-regions, minor changes have been adapted to the above procedures based on the similar idea in \cite{gao2005image}. The key is to divide the values of the resulted evolving curve $\varphi$ into two parts: inside ($\varphi \ >0$) and outside ($\varphi \ <0$). Then the AMA segmentation algorithm is applied again for the inside and outside regions, respectively. This set of segmentation for the inside and outside regions will be called second level segmentation and the whole process will be called two-level segmentation. This procedure can be repeated a few times depending on the number of interested regions, for example, three-level segmentation.

\section{Numerical results}
\label{Sec-results}

In this section we present some results obtained using our AMA segmentation method and compare them with those obtained by standard finite difference schemes (FDS). 
If not stated otherwise, the following values are used for parameters in \eqref{pdemodel}: $\lambda_1=\lambda_2=1$, $\nu=0$, $\epsilon=1$. Two values are used for $\mu$ in the computations for different images, one is $\mu=0.0001$ and the other is $\mu=0.01$. Larger $\mu$ value produces smoother boundary while smaller value produces faster segmentation (see \cite{getreuer2012chan}).
For the diffusion term defined in \eqref{defD}, a regularized form, $\DD = \frac{\mu}{1 +\ |\nabla\varphi|}$, is used in the computation. 

The following notations are also used to describe the results. \textit{Iterations} is the number of iterations needed for the evolving curve to converge to the solution. \textit{Adaptive-time} is the time needed for mesh adaptation in the stage of AMA representation. \textit{Total-time} is the total time needed to finish the computation including mesh adaptation. \textit{Adaptive-time} and \textit{Total-time} are both measured using \textit{tic} and \textit{toc} functions in MATLAB. \textit{dt} is the time step used in the discretization of time domain, and \textit{sd} is the sample density used to represent the original image with anisotropic mesh.\\

\begin{exam}
\label{ex-1}
For the first example, we demonstrate the effectiveness of AMA segmentation method on an image with 10 circles which we call image Circles (taken from \cite{getreuer2012chan}). The image has resolution of $1600 \times 1600$ and has only one region of interests, that is, the circles. We compare two different metric tensors, $\MM_{DMP}$ and $\MM_{aniso}$, for AMA segmentation, and denote the corresponding results as $\MM_{DMP}$ segmentation and $\MM_{aniso}$ segmentation, respectively. Note that the metric tensors are used only in the AMA representation stage. Once the AMA mesh is generated according to the corresponding metric tensors in AMA representation stage, the mesh will be fixed and provided as the initial mesh for solving the PDE model. The results are also compared with those obtained using finite different scheme (FDS).  dt=1000 and $\mu=0.0001$ are used in the computations for this example.

Fig. \ref{fig:fig-Circles}(a) shows the original image Circles. Fig. \ref{fig:fig-Circles}(b) displays the evolving curve of $\MM_{DMP}$ segmentation, and Fig. \ref{fig:fig-Circles}(c) shows the corresponding results. The computational times are \textit{Adaptive-time}=11.5s and \textit{Total-time}=12.5s. Fig. \ref{fig:fig-Circles}(d) display the results of $\MM_{aniso}$ segmentation. The computational times are \textit{Adaptive-time}=11.3s and \textit{Total-time}=12.2s. For both $\MM_{DMP}$ segmentation and $\MM_{aniso}$ segmentation, two iterations are sufficient for the evolving curve to converge. On the other hand, FDS needs many more iterations to achieve reasonably good results. The segmentation obtained using FDS after 50 iterations is shown in Fig. \ref{fig:fig-Circles}(e), and the computational time is \textit{Total-time}=53s.

\begin{figure}[hbt!]
	\begin{center}
		\begin{subfigure}[normal]{0.2\textwidth}
			\includegraphics[scale=0.45]{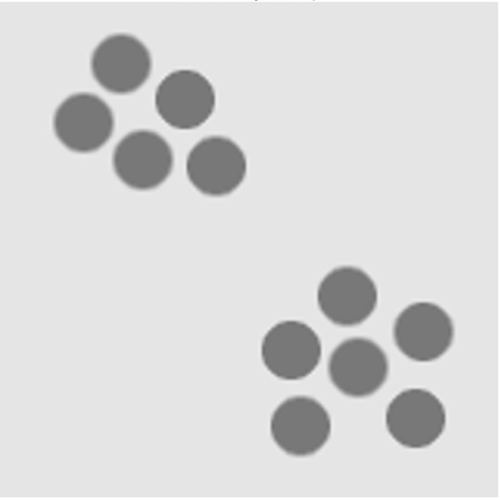}
			\caption{}
			
		\end{subfigure}
		\hspace{5mm}
		\begin{subfigure}[normal]{0.2\textwidth}
			\includegraphics[scale=0.45]{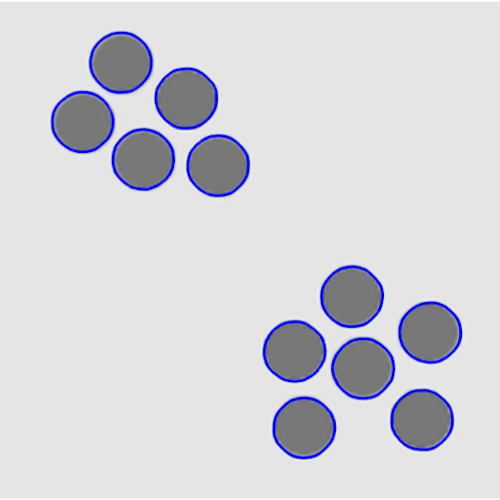}
			\caption{}
			
		\end{subfigure}
		
		\vspace{2mm}
		
		\begin{subfigure}[normal]{0.2\textwidth}
			\includegraphics[scale=0.45]{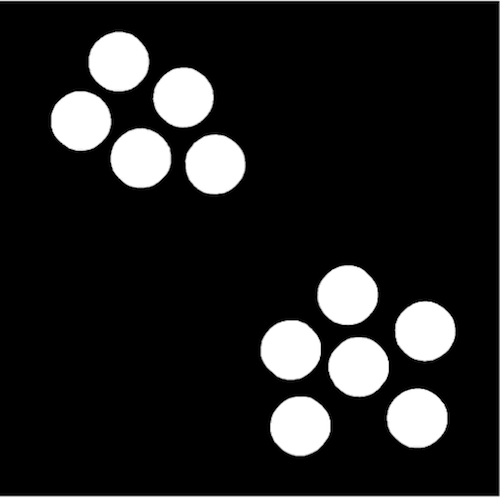}
			\caption{$\MM_{DMP}$ segmentation}
		\end{subfigure}
		\hspace{5mm}
		\begin{subfigure}[normal]{0.2\textwidth}
			\includegraphics[scale=0.45]{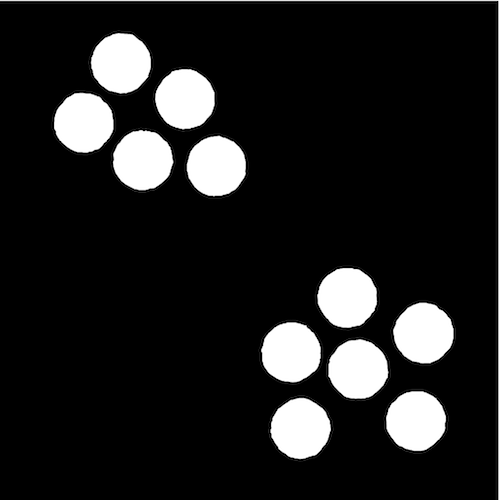}
			\caption{$\MM_{aniso}$ segmentation}
		\end{subfigure}
		\hspace{5mm}
		\begin{subfigure}[normal]{0.2\textwidth}
			\includegraphics[scale=0.45]{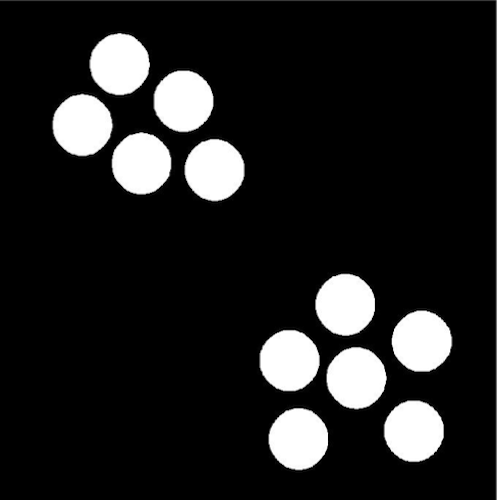}
			\caption{FDS}
		
		\end{subfigure}
		\caption{Example \ref{ex-1}. One-level AMA Segmentation of image Circles (taken from \cite{getreuer2012chan}) with resolution $1600 \times 1600$, $dt$=1000, $sd$=0.001, $\mu=0.0001$. (a) original image, (b) evolved curve, (c) $\MM_{DMP}$ segmentation, \textit{Iterations}=2, \textit{Total-time}=12.5s, \textit{Adaptive-time}=11.5s, (d) $\MM_{aniso}$ segmentation, \textit{Iterations}=2, \textit{Total-time}=12.2s, \textit{Adaptive-time}=11.3s, (e) FDS, \textit{Iterations}=50, \textit{Total-time}=53s.}
		\label{fig:fig-Circles}
		\end{center}
	
	\end{figure}

The results clearly show that our AMA segmentation improves computational efficiency significantly over the traditional FDS. Furthermore, the results obtained using metric tensors $\MM_{DMP}$ and $\MM_{aniso}$ in AMA segmentation are comparable. 

Comparing the results in Fig. \ref{fig:fig-Circles}(c), Fig. \ref{fig:fig-Circles}(d), and Fig. \ref{fig:fig-Circles}(e), we observe that $\MM_{DMP}$ segmentation has smoother edges for the segmented regions than $\MM_{aniso}$ segmentation and FDS. Figure \ref{fig-aniso-and-DMP} shows the comparison between $\MM_{DMP}$ segmentation and $\MM_{aniso}$ segmentation in more details, where $N_e$ denotes the number of elements in the mesh. Figure \ref{fig-aniso-and-DMP}(a) and \ref{fig-aniso-and-DMP}(b) show that more elements concentrate around the edges in $\MM_{DMP}$ mesh than in $\MM_{aniso}$ mesh.

\begin{figure}[hbt!]
\begin{center}
\hspace{2mm}
\begin{minipage}[b]{2in}
\includegraphics[width=2in]{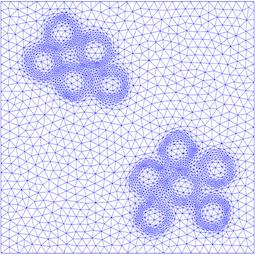}
\centerline{(a) $\MM_{aniso}$ mesh, $N_e=6036$}
\end{minipage}
\hspace{2mm}
\begin{minipage}[b]{2in}
\includegraphics[width=2in]{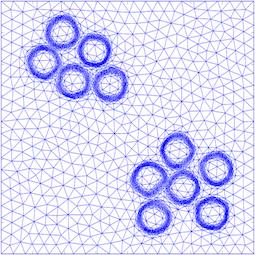}
\centerline{(b) $\MM_{DMP}$ mesh, $N_e=6090$}
\end{minipage}

\vspace{5mm}
\hspace{2mm}
\begin{minipage}[b]{2in}
\includegraphics[width=2in]{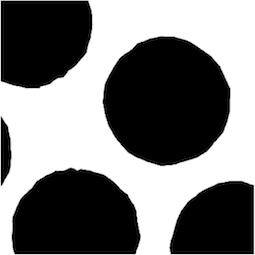}
\centerline{(c) $\MM_{aniso}$ segmentation}
\end{minipage}
\hspace{2mm}
\begin{minipage}[b]{2in}
\includegraphics[width=2in]{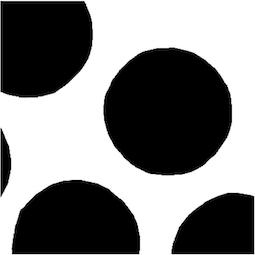}
\centerline{(d) $\MM_{DMP}$ segmentation}
\end{minipage}
\caption{Example \ref{ex-1}. AMA Segmentation using different metric tensors for image Circles. (a) $\MM_{aniso}$ mesh,  (b) $\MM_{DMP}$ mesh, (c) enlarged view of $\MM_{aniso}$ segmentation, (d) enlarged view of $\MM_{DMP}$ segmentation.}
\label{fig-aniso-and-DMP}
\end{center}
\end{figure}

On the other hand, mesh adaptation in $\MM_{DMP}$ segmentation takes a little more computational time than $\MM_{aniso}$ segmentation. Overall, the two AMA representation methods are comparable. In our computation, we choose the $\MM_{aniso}$ segmentation which provides good results most of the time. If smoother edges are desired, we will choose $\MM_{DMP}$ segmentation.
\end{exam}

\begin{exam}
\label{ex-2}
In this example, we perform segmentation on more complex real life images including image Bacteria (taken from \cite{Tatyana2019Acute}) with resolution $1024 \times 1024$ and image Bear (taken from \cite{Ricardo2015Bow}) with resolution $2048 \times 2048$. Figures \ref{fig:fig-bactria} and \ref{fig:fig-bear} show the results for image Bacteria and image Bear, respectively. The results are similar to those from image Circle. 

For image Bacteria, FDS does not converge when using large time step, therefore, smaller time step $dt=0.5$ is used in the computations. It takes 400 iterations and 186s total computational time to obtain a reasonably good result using FDS. While, $\MM_{aniso}$ segmentation only needs 3 iteration and 7.1s computational time to obtain a better result than FDS. $\MM_{DMP}$ segmentation only takes 1 iteration to converge, but the result is not as good as that from $\MM_{aniso}$ segmentation. 

For image Bear, $dt=1000$ works for both FDS and AMA segmentation algorithms. Again, FDS needs more iterations to converge and thus takes more computational time. The segmentation results are comparable for this case.

\begin{figure}[H]
	\begin{center}
		\begin{subfigure}[normal]{0.2\textwidth}
			\includegraphics[scale=0.425]{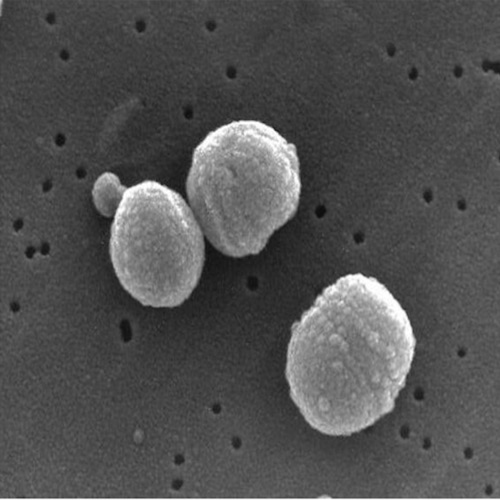}
			\caption{}
		\end{subfigure}
		\hspace{5mm}
		\begin{subfigure}[normal]{0.2\textwidth}
			\includegraphics[scale=0.425]{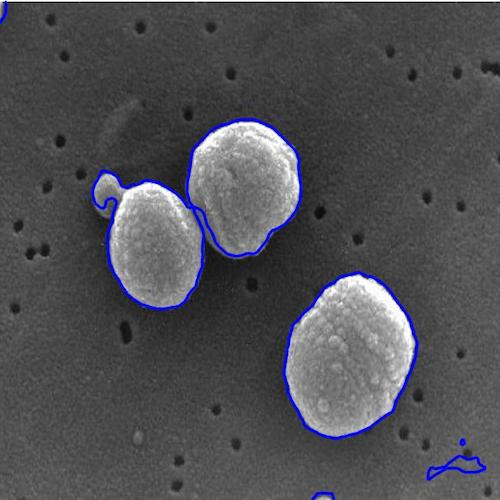}
			\caption{}
		\end{subfigure}
		
		\begin{subfigure}[normal]{0.2\textwidth}
			\includegraphics[scale=0.425]{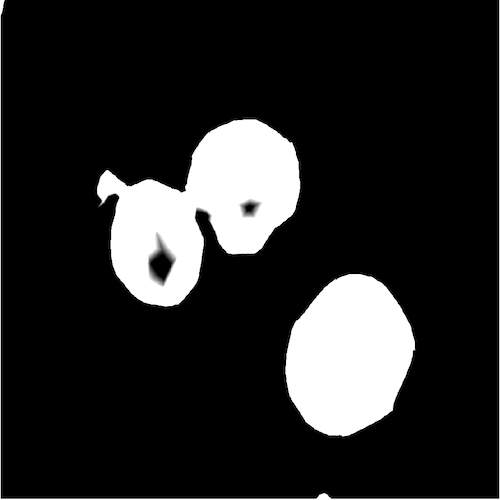}
			\caption{$\MM_{DMP}$}
		\end{subfigure}
		\hspace{5mm}
		\begin{subfigure}[normal]{0.2\textwidth}
			\includegraphics[scale=0.425]{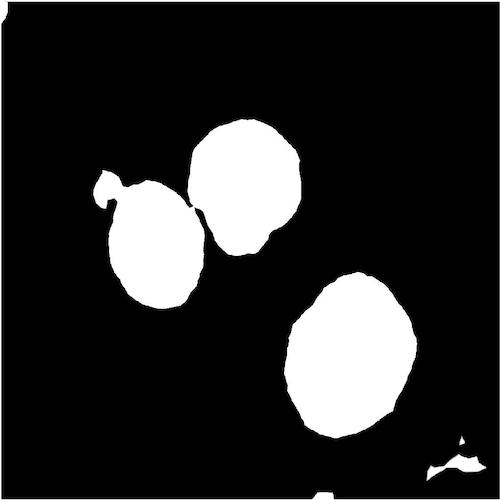}
			\caption{$\MM_{aniso}$}
		\end{subfigure}
		\hspace{5mm}
		\begin{subfigure}[normal]{0.2\textwidth}
			\includegraphics[scale=0.425]{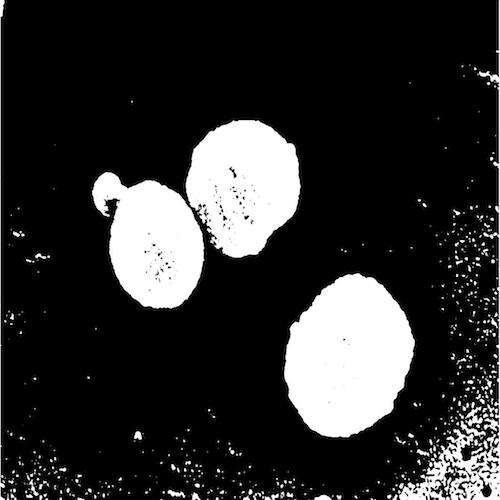}
			\caption{FDS}
		\end{subfigure}
		\caption{Example \ref{ex-2}. One-level AMA Segmentation of image Bacteria (taken from \cite{Tatyana2019Acute}) with resolution $1024 \times 1024$, $dt$=1000, $sd$=0.002, $\mu=0.0001$. (a) original image, (b) evolved curve, c) $\MM_{DMP}$ segmentation, \textit{Iterations}=1, \textit{Total-time}=5.4s, \textit{Adaptive-time}=5.1s, (d) $\MM_{aniso}$ segmentation, \textit{Iterations}=3, \textit{Total-time}=7.1s, \textit{Adaptive-time}=6.2s, (e) FDS, \textit{Iterations}=400, \textit{Total-time}=186s, dt=0.5. }
		\label{fig:fig-bactria}
		\end{center}
	
	\end{figure}

 \begin{figure}[htb]
	\begin{center}
		\begin{subfigure}[normal]{0.2\textwidth}
			\includegraphics[scale=0.425]{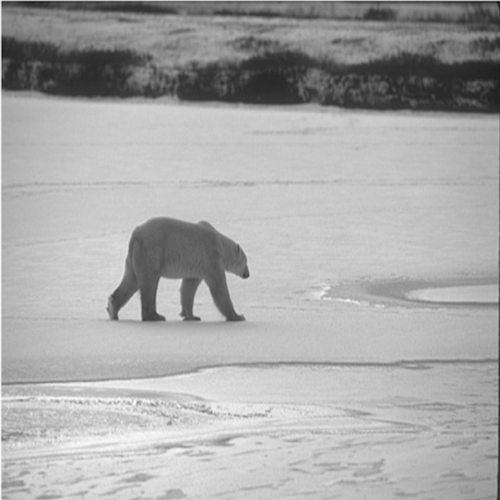}
			\caption{}
		\end{subfigure}
		\hspace{5mm}
		\begin{subfigure}[normal]{0.2\textwidth}
			\includegraphics[scale=0.425]{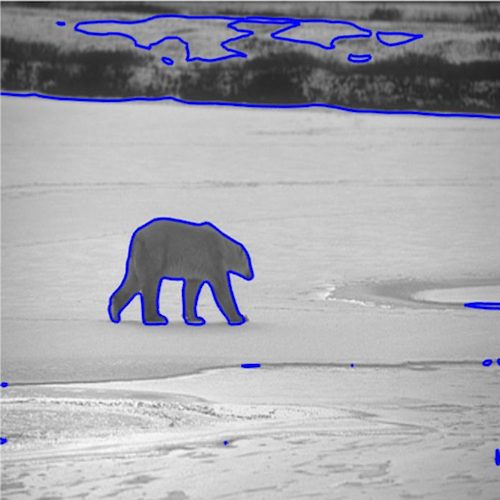}
			\caption{}
		\end{subfigure}
		
		
		\begin{subfigure}[normal]{0.2\textwidth}
			\includegraphics[scale=0.425]{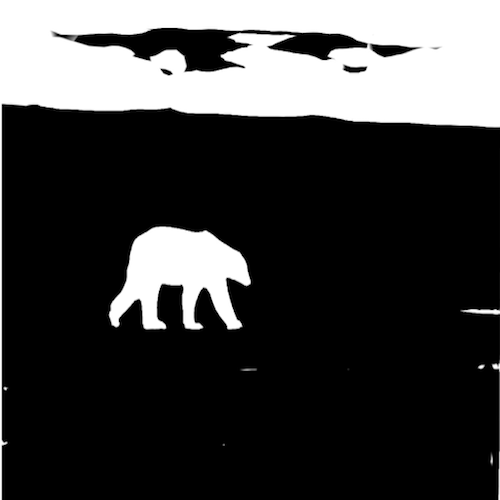}
			\caption{$\MM_{DMP}$}
		\end{subfigure}
		\hspace{5mm}
		\begin{subfigure}[normal]{0.2\textwidth}
			\includegraphics[scale=0.425]{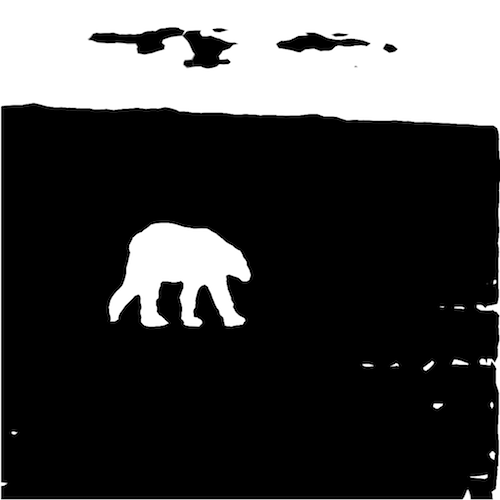}
			\caption{$\MM_{aniso}$}
		\end{subfigure}
		\hspace{5mm}
		\begin{subfigure}[normal]{0.2\textwidth}
			\includegraphics[scale=0.425]{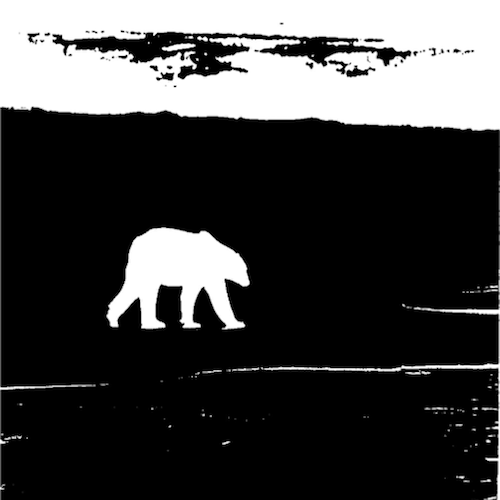}
			\caption{FDS}
		\end{subfigure}
		\caption{Example \ref{ex-2}. One-level AMA Segmentation of image Bear (taken from \cite{Ricardo2015Bow}) with resolution $2048 \times 2048$, $dt$=1000, $sd$=0.001, $\mu=0.01$, (a) original image, (b) evolved curve, (c) $\MM_{DMP}$ segmentation, \textit{Iterations}=3, \textit{Total-time}=24.5s, \textit{Adaptive-time}=21.1s, (d) $\MM_{aniso}$ segmentation, \textit{Iterations}=3, \textit{Total-time}=20.6s, \textit{Adaptive-time}=22.9s, (e) FDS, \textit{Iterations}=30, \textit{Total-time}=62s. }
		\label{fig:fig-bear}

	\end{center}
	
    \end{figure}	
	
Overall, the results confirm that our AMA segmentation algorithm works both faster and better than FDS.

\end{exam}

\begin{exam}
\label{ex-3}
In this example, we perform segmentation on an noisy image denoted as image Noise (see Figure \ref{fig:fig-NoisyImage}(a)). The resolution of the image is $2048 \times 2048$. $dt=1000$ is used for both FDS and AMA algorithms. The results are shown in Figure \ref{fig:fig-NoisyImage}.

The evolved curve of AMA segmentation shown in Figure \ref{fig:fig-NoisyImage}(b) was based on the $\MM_{DMP}$ segmentation. The segmentation result is shown in Figure \ref{fig:fig-NoisyImage}(c). The computational time is \textit{Total-time}=32.5s. $\MM_{aniso}$ segmentation takes a little less time with \textit{Total-time}=27.7s and the result is displayed in Figure \ref{fig:fig-NoisyImage}(d). Figure \ref{fig:fig-NoisyImage}(e) shows the result from FDS after \textit{Total-time}=250s, and the solution still does not converge. In fact, FDS does not converge even after more iterations or with different initialization curves. However, after downscaling the image to lower resolution, for example, $512 \times 512$, FDS provides a good representation after 500 iterations. 


\begin{figure}[H]
	\begin{center}
		\begin{subfigure}[normal]{0.2\textwidth}
			\includegraphics[scale=0.45]{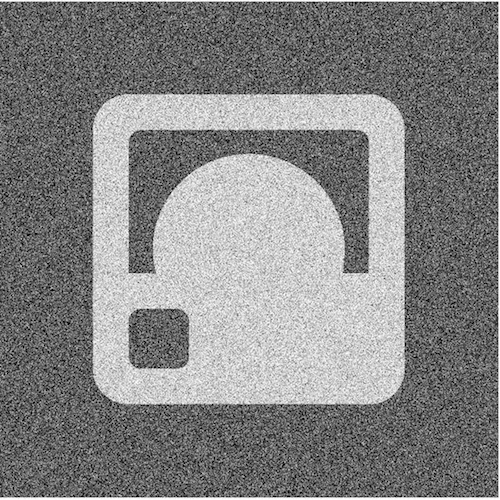}
			\caption{}
		\end{subfigure}
	     \hspace{5mm}
		\begin{subfigure}[normal]{0.2\textwidth}
			\includegraphics[scale=0.45]{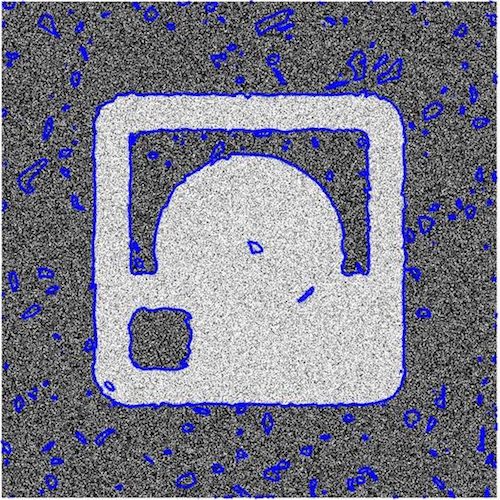}
			\caption{}
		\end{subfigure}
		
		\vspace{1mm}
		
		\begin{subfigure}[normal]{0.2\textwidth}
			\includegraphics[scale=0.45]{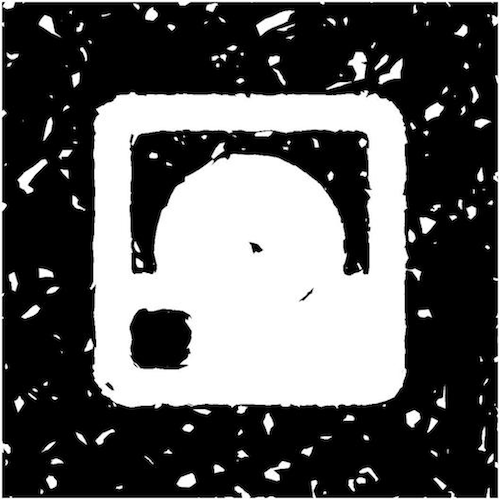}
			\caption{$\MM_{DMP}$}
		\end{subfigure}
		\hspace{5mm}
		\begin{subfigure}[normal]{0.2\textwidth}
			\includegraphics[scale=0.45]{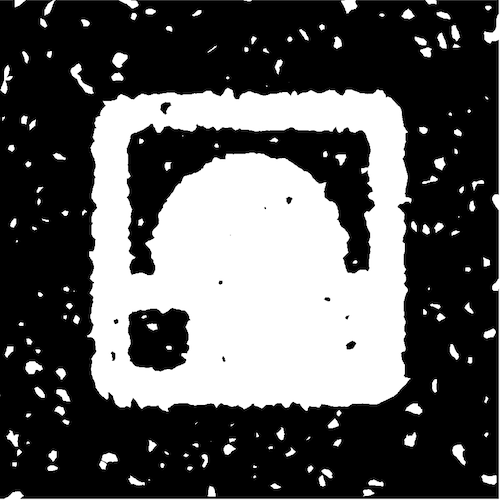}
			\caption{$\MM_{aniso}$}
		\end{subfigure}
		\hspace{5mm}
		\begin{subfigure}[normal]{0.2\textwidth}
			\includegraphics[scale=0.45]{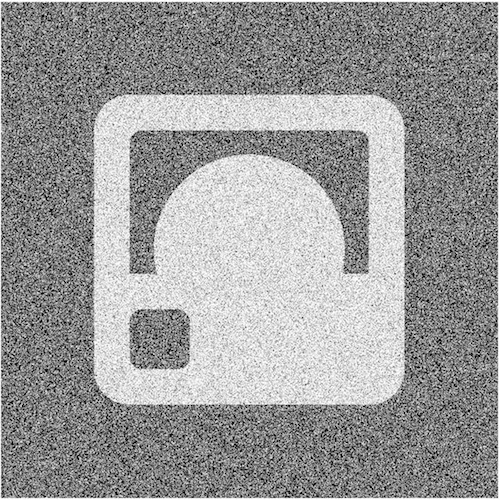}
			\caption{FDS}
		\end{subfigure}
		\caption{Example \ref{ex-3}. One-level AMA Segmentation of image Noise with resolution $2048 \times 2048$, $dt$=1000, $sd$=0.002, $\mu=0.01$. (a) original image, (b) evolved curve, (c) $\MM_{DMP}$ segmentation, \textit{Iterations}=4,  \textit{Total-time}=32.5s, \textit{Adaptive-time}=22.4s, (d) $\MM_{aniso}$ segmentation, \textit{Iterations}=4,  \textit{Total-time}=27.7s, \textit{Adaptive-time}=22.8s, (e) FDS, \textit{Iterations}=100s, \textit{Total-time}=250s.}
		\label{fig:fig-NoisyImage}

	\end{center}
	
    \end{figure}

The results demonstrate that our AMA segmentation works much better than FDS for segmentation of high resolution noisy images. While the results from $\MM_{DMP}$ segmentation (Figure \ref{fig:fig-NoisyImage}(c)) and $\MM_{aniso}$ segmentation (Figure \ref{fig:fig-NoisyImage}(d)) are comparable, $\MM_{DMP}$ segmentation performs better in terms of denoising and smoothing of the noisy image.  

\end{exam}

\begin{exam}
\label{ex-4}
To extent our exploration, we consider images with multi-regional segments in this example. Two images are chosen for this study: image Duck (Figure \ref{fig:Duck}(a), taken from \cite{BryanPb}) and image Rings (Figure \ref{fig:Rings}(a), taken from \cite{Ernest1925Concentric}), both are of resolution $1080 \times 1080$. The images have at least two or more regions besides the background. In order to obtain the segmentation for multiple regions, we adapt our AMA scheme to perform multi-level segmentations. Metric tensor $\MM_{aniso}$ is used for this example. The results are shown in Figures \ref{fig:Duck} and \ref{fig:Rings}, respectively.

Two-level AMA segmentation is performed for image Duck (Figure \ref{fig:Duck}(a)). The time step is the same for both levels, that is, $dt=1000$. Two iterations are sufficient for the first level, and six iterations are used for the second level. The total computational time is \textit{Total-time}=15.1s. The results are shown in Figures \ref{fig:Duck}.

 
 \begin{figure}[H]
\begin{center}
\begin{minipage}[b]{1.5in}
\includegraphics[width=1.5in]{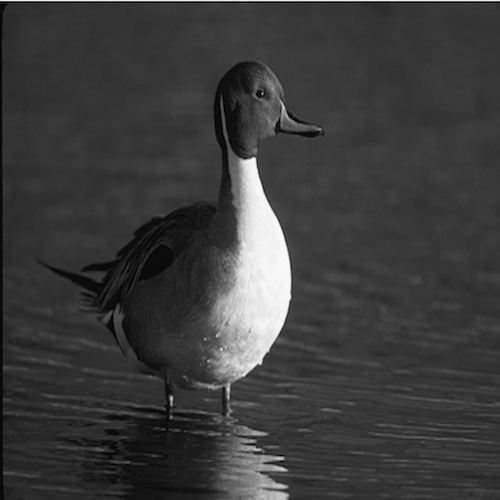}
\centerline{(a)}
\end{minipage}
\hspace{1mm}
\begin{minipage}[b]{1.5in}
\includegraphics[width=1.5in]{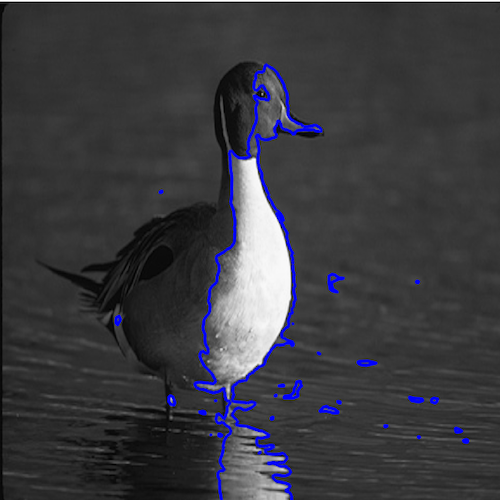}
\centerline{(b)}
\end{minipage}
\hspace{1mm}
\begin{minipage}[b]{1.5in}
\includegraphics[width=1.5in]{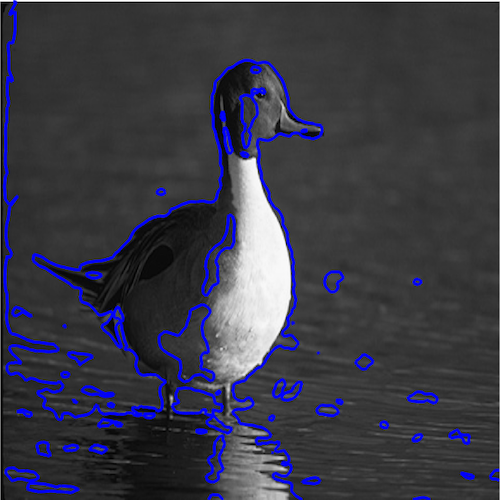}
\centerline{(c)}
\end{minipage}

\vspace{2mm}

\begin{minipage}[b]{1.5in}
\includegraphics[width=1.5in]{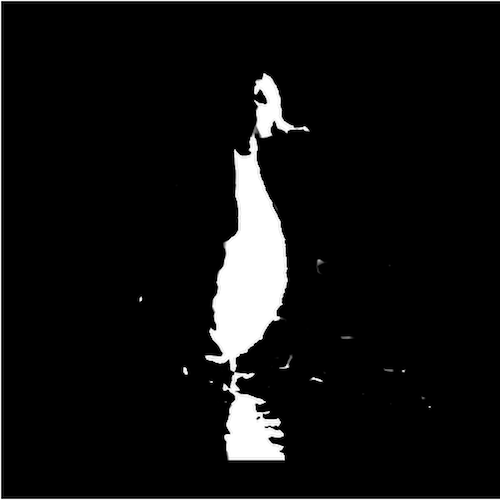}
\centerline{(d) first-level}
\end{minipage}
\hspace{1mm}
\begin{minipage}[b]{1.5in}
\includegraphics[width=1.5in]{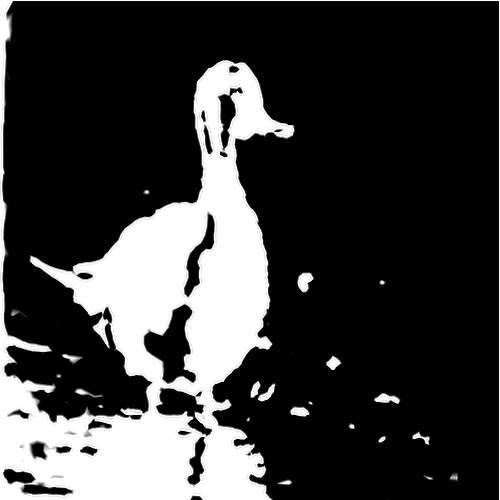}
\centerline{(e) second-level}
\end{minipage}
\caption{Example \ref{ex-4}. Two-level AMA segmentation of image Duck (taken from \cite{BryanPb}) with resolution $1080 \times 1080$, $dt$=1000 (for both levels), $sd$=0.004, $\mu=0.01$, \textit{Iterations}=2 (first level), 6 (second level), \textit{Total-time}=15.1s, \textit{Adaptive-time}=8.6s. (a) original image, (b) evolved curve for the first level, (c) evolved curve for the second level, (d) first-level segmentation, (e) second-level segmentation.}
\label{fig:Duck}
\end{center}
\end{figure}

The first-level AMA segmentation only provides results for the right half of the duck, see Figures \ref{fig:Duck}(b) and (d). Then the second-level AMA segmentation provides results for the left half of the duck, see Figure \ref{fig:Duck}(c). Figure \ref{fig:Duck}(e) shows the result after the second-level segmentation where both regions are combined together. As can be seen from Figure \ref{fig:Duck}(e), the middle part of the duck was treated as background due to its gray value being very close to the background.

For image Rings (Figure \ref{fig:Rings}(a)), the regions are defined by the 6 curves (circles). Three-level AMA segmentation is performed to catch all the curves, and $dt=1000$ is used in all levels. Two iterations are sufficient for the first level, four iterations are needed for the second level, and seven iterations are used for the third level. The total computational time is \textit{Total-time}=57s. The results are shown in Figure \ref{fig:Rings}.


\begin{figure}[H]
\begin{center}
\begin{minipage}[b]{1.5in}
\includegraphics[width=1.5in]{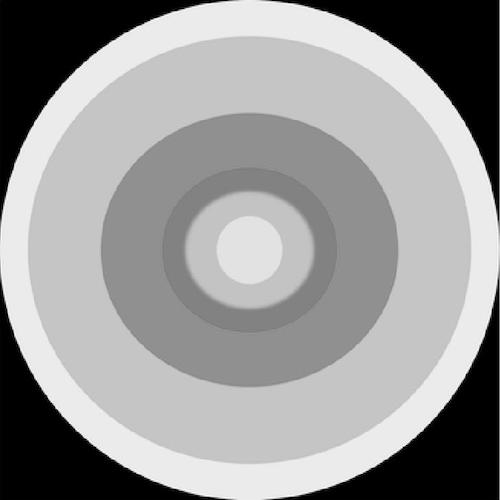}
\centerline{(a)}
\end{minipage}
\hspace{1mm}
\begin{minipage}[b]{1.5in}
\includegraphics[width=1.5in]{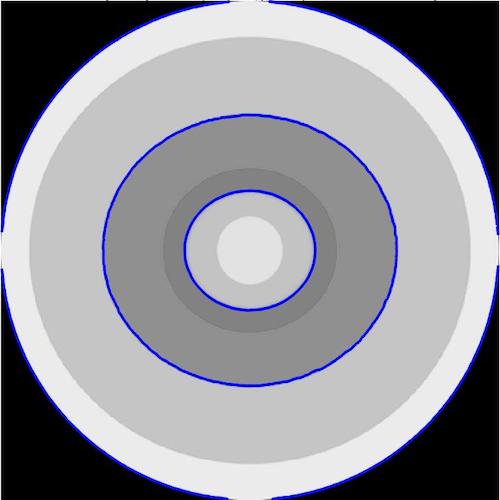}
\centerline{(b)}
\end{minipage}
\hspace{1mm}
\begin{minipage}[b]{1.5in}
\includegraphics[width=1.5in]{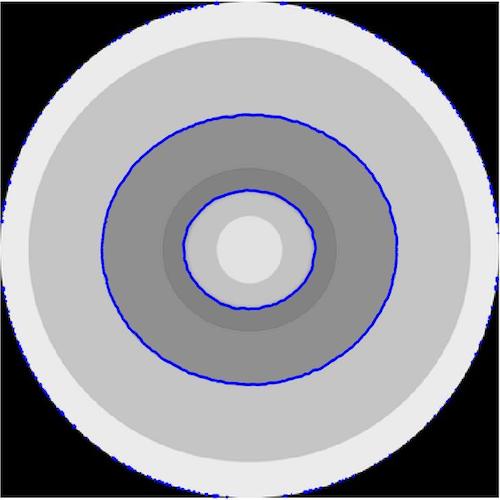}
\centerline{(c)}
\end{minipage}

\vspace{2mm}

\begin{minipage}[b]{1.5in}
\includegraphics[width=1.5in]{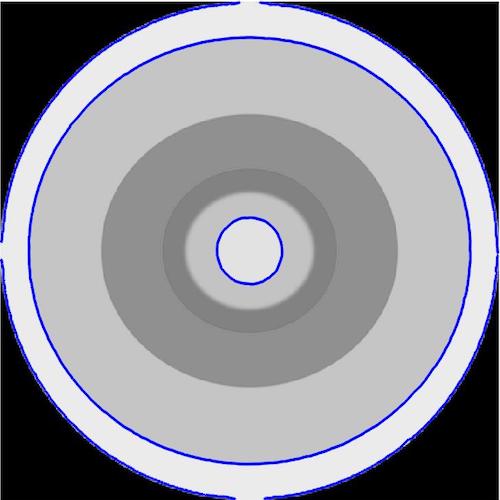}
\centerline{(d)}
\end{minipage}
\hspace{1mm}
\begin{minipage}[b]{1.5in}
\includegraphics[width=1.5in]{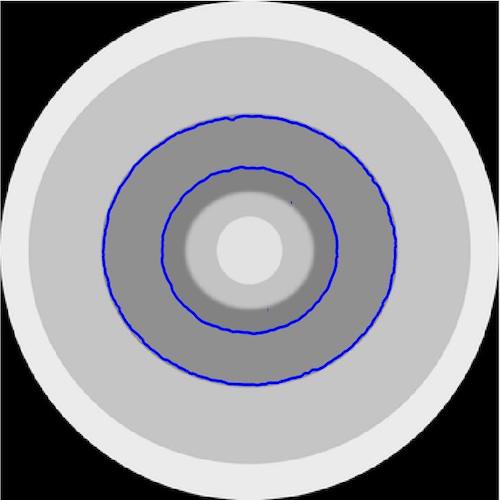}
\centerline{(e)}
\end{minipage}
\hspace{1mm}
\begin{minipage}[b]{1.5in}
\includegraphics[width=1.5in]{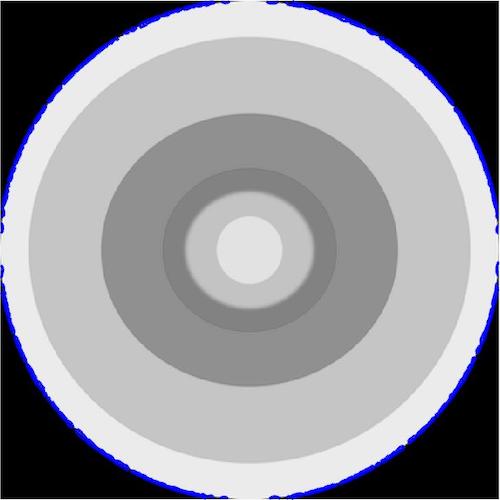}
\centerline{(f)}
\end{minipage}
 
\caption{Example \ref{ex-4}. Three-level AMA Segmentation of image Rings (taken from \cite{Ernest1925Concentric}) with resolution $1080 \times 1080$, $dt$=1000, $sd$=0.0085, $\mu=0.0001$, \textit{Iterations}=2 (first level), 4 (second level), 7 (third level), \textit{Total-time}=57s, \textit{Adaptive-time}=16.6s. (a) original image, (b) evolved curve for the first level, (c) evolved curve for the second level (inside region), (d) evolved curve for the second level (outside region), (e) evolved curve for the third level (inside Region), (f) evolved curve for the third level (outside region).}
	\label{fig:Rings}
\end{center}
\end{figure}

The first-level segmentation provides result for two evolved curves as shown in Figures \ref{fig:Rings}(b). The second-level segmentation shows two additional evolved curves from outside region (Figure \ref{fig:Rings}(d)), while evolving inside does not provide any new curve. The third-level segmentation provides two additional evolved curves, one from inside region (Figure \ref{fig:Rings}(e)) and one from outside region (Figure \ref{fig:Rings}(f)). After three levels, we are able to catch all the circles (curves).

\end{exam}
		
\section{Conclusions and comments}
\label{Sec-con}
Partial differential equation (PDE) methods have become a popular method in image segmentation due to its solid mathematical foundation and numerical stability. However, their numerical computations are usually time intensive. Anisotropic mesh adaptation has been successfully used in solving PDEs by many researchers. In this paper, we have introduced a framework of anisotropic mesh adaptation methods (AMA) for image segmentation with the main purpose of improving computational efficiency as well as accuracy. 

In our AMA segmentation method, we first represent the image with fewer points (pixels) than the original image using an anisotropic triangular mesh. Then we use the corresponding mesh along with finite element method to solve the PDE model \eqref{pdemodel} and \eqref{pdebc} derived from the minimization of the Mumford-Shah functional. 

We have applied the AMA segmentation algorithm to a few test images, and compared the results with those obtained using traditional finite difference schemes (FDS). Images in Examples \ref{ex-1} and \ref{ex-2} do not have noise and the main features are in one region. Thus one-level AMA segmentation has been performed, and the results are better than those obtained using FDS. Our AMA segmentation scheme is also much faster than FDS.  

We have also considered segmentation for a noisy image in Example \ref{ex-3}. Traditional FDS does not converge unless the image is downscaled to lower resolution. However, our AMA segmentation converges after 4 iterations and takes less than 35s to complete the task. Furthermore, $\MM_{DMP}$ segmentation deals with noise better than $\MM_{aniso}$ segmentation.  

Our AMA segmentation algorithm is also adapted to perform multi-level segmentation for images containing multiple regions. The results from Example \ref{ex-4} demonstrate the effectiveness of the multi-level AMA segmentation. 

In summary, we have introduced an AMA segmentation method that has significant advantage over traditional finite difference schemes in terms of both computational efficiency and segmentation results. AMA segmentation can also deal well with noisy images and images with multiple regions. The method has potential applications in other areas such as medical image segmentation.

\end{document}